\documentclass[sigconf]{acmart}
\acmSubmissionID{1574}
\usepackage{multirow}
\usepackage{cleveref}
\usepackage{natbib}
\usepackage{marvosym}
\usepackage[utf8]{inputenc}
\AtBeginDocument{%
  }

\setcopyright{acmlicensed}
\copyrightyear{2018}
\acmYear{2018}
\acmDOI{XXXXXXX.XXXXXXX}
\acmConference[Conference acronym 'XX]{Make sure to enter the correct
  conference title from your rights confirmation email}{June 03--05,
  2018}{Woodstock, NY}

\setcitestyle{square, comma} 

\begin{document}

\title{Mirage2Matter: A Physically Grounded Gaussian World Model from Video}


\author{Zhengqing Gao$^{*}$, Ziwen Li$^{*}$, Xin Wang, Jiaxin Huang, Zhenyang Ren, Mingkai Shao, Hanlue Zhang, Tianyu Huang, Yongkang Cheng, Yandong Guo, Runqi Lin, Yuanyuan Wang, Tongliang Liu, Kun Zhang, Mingming Gong$^{\textsuperscript{\Letter}}$}
\affiliation{
  \institution{MBZUAI, AI$^2$Robotics, The University of Sydney, Carnegie Mellon University, The University of Melbourne}
  \country{UAE,  China,  Australia,  USA}
  }

\renewcommand{\shortauthors}{Anonymous Authors}

\begin{abstract}

The scalability of embodied intelligence is fundamentally constrained by the scarcity of real-world interaction data. 
While simulation platforms provide a promising alternative, existing approaches often exhibit a substantial visual and physical gap to real environments and require specialized hardware for data collection, limiting their practicality at scale.
To bridge the simulation-to-real gap, we introduce \textbf{Mirage2Matter}, a physically grounded Gaussian world model that generates high-fidelity embodied training data from only multi-view videos.
Specifically, our approach reconstructs real-world environments into a photorealistic scene representation using 3D Gaussian Splatting (3DGS), seamlessly capturing fine-grained geometry and appearance from video. 
We then leverage generative models to recover a physically realistic representation and integrate it into a simulation environment via a precision calibration target, enabling accurate scale alignment between the reconstructed scene and the real world. 
By coupling photorealistic reconstruction with physically grounded simulation, our design enables a unified, editable, and reality-consistent visual world model.
Vision-Language-Action (VLA) models trained with our Mirage2Matter framework achieve strong zero-shot performance on downstream tasks, even matching models trained on real-world data and providing a practical path toward scalable embodied intelligence.


\end{abstract}


\begin{teaserfigure}
  \includegraphics[width=\textwidth]{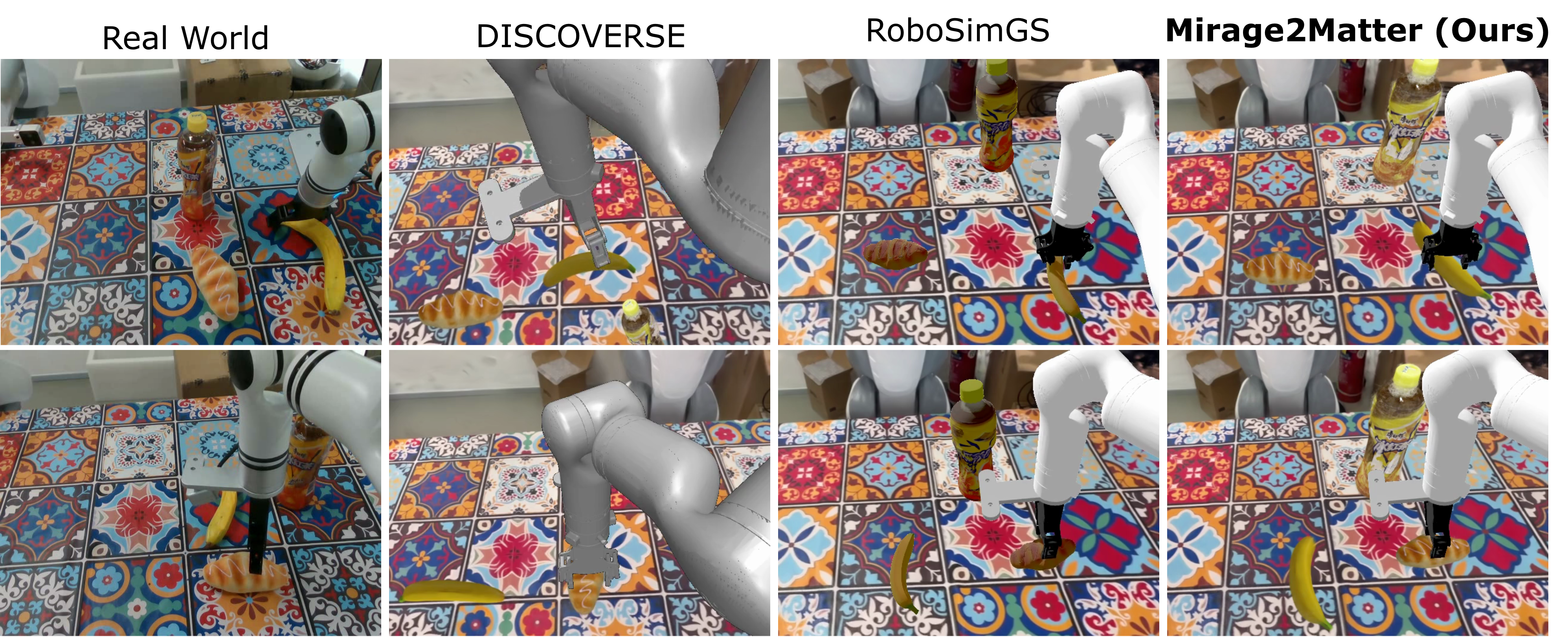}
  \caption{
\textbf{From photorealistic simulation to real-world execution.}
We construct a physics-grounded simulator using 3D Gaussian representations
that faithfully capture both the appearance and geometry of real environments and objects.
Policies trained entirely in this simulator can be directly transferred to the physical robot
and execute the same manipulation tasks in the real world in a zero-shot manner,
without any additional fine-tuning.
}
  \label{fig:teaser}
\end{teaserfigure}


\maketitle

\section{Introduction}
Despite remarkable progress in artificial intelligence (AI), progress toward embodied intelligence has remained limited due to the complexity of real-world environments and the constraints imposed by physical laws.
Recently, a growing body of work has shown that vision-language-action (VLA) models can learn from real-world data to perform physical manipulation and interaction tasks, offering a promising path toward embodied AI \cite{visionary, interndataA1, vlabench,black2410pi0,bu2025univla,kim2024openvla,chi2025diffusion}.
However, current VLA training paradigms rely heavily on high-quality real-world data, which is expensive and time-consuming to collect, thereby hindering scalability in practice.
In light of this, previous works have leveraged visual world models to generate large-scale, cost-effective data for training VLA models.

Current simulation techniques \cite{hofer2021sim2real,li2021igibson,ehsani2021manipulathor} can generally be divided into two categories, reconstruction-based and generative-based world models. Conventional reconstruction-based 3D simulation environments, such as AI2thor \cite{ai2thor}, Habitat series \cite{habitat, habitat2, habitat_third} and SAPIEN \cite{sapien}, provide basic rendering pipelines for visual navigation and manipulation. However, these systems rely on manually authored assets and synthetic textures, resulting in a noticeable domain gap relative to real-world imagery.
Later extensions introduced 3DGS to enable photorealistic rendering faithful to real-world appearance and incorporated procedural layout variations to narrow this gap, including DISCOVERSE \cite{discoverse}, RoboGSim \cite{robogsim}, Re3Sim \cite{re3sim}, 3DGSim \cite{3dgsim}, Polaris \cite{polaris} and  GSWorld \cite{gsworld}, PhysSplat \cite{physsplat}, Giga \cite{giga}, SplatSim \cite{splatsim}, and ScanSim \cite{scansim}. 
Despite improved visual realism, these approaches often depend on expensive sensors, precise robot calibration, or depth measurements, which hinder personalized robot learning from ordinary videos.
Moreover, they lack a mechanism to capture physical laws and thus require additional scene setup and task-specific solvers, which further reduces real-world practicality.

On the other hand, generative-based world models, such as CtrlWorld \cite{ctrlworld}, EmbodiedGen \cite{embodiedgen}, PhysGen3D \cite{physgen3d}, OmniPhysGS \cite{omniphysgs}, LucidSim \cite{lucidsim} and DreamGen \cite{dreamgen}, offer an automated pipeline that synthesizes 3D assets from multi-view videos and assembles them into a simulated environment. In addition, generative models can predict physical properties, enabling more realistic and dynamics-consistent manipulation and interaction behaviors. Unfortunately, substantial mismatches often remain between generated assets and their real-world counterparts, introducing harmful noise into the synthesized data. These discrepancies degrade environmental fidelity and consequently limit the utility of the resulting simulation. As a result, current world models still struggle to simultaneously achieve high-fidelity and physical grounding, producing simulated data with a simulation-to-real gap that undermines zero-shot deployment of VLA models trained on it in the real world.

To this end, we propose \textit{\textbf{Mirage2Matter}}, a Gaussian world model that combines photorealistic rendering with physically grounded interaction dynamics to enable reality-consistent data generation. Specifically, our pipeline reconstructs high-quality 3DGS representations of both target environments and manipulable objects from ordinary multi-view videos, capturing fine-grained geometry and appearance to build photorealistic scenes. To endow manipulable objects with physically grounded properties, we leverage generative models to efficiently produce collision-ready geometric representations, providing realistic interaction feedback. Using the robot’s egocentric camera as an anchor, we introduce a calibration-and-alignment pipeline that dynamically preserves consistency between visual and physical representations, ensuring that the simulated observations continuously match their real-world counterparts. 

In addition, our simulation platform supports the inserting of new assets, the editing of existing assets within the reconstructed environment, and the faithful reproduction of real-world lighting conditions, further reducing the visual gap between simulation and reality. Overall, our approach enables the low-cost construction of visually faithful world models with grounded interaction dynamics directly from real environments, without requiring specialized hardware or extensive manual configuration. Extensive experiments show that \textit{\textbf{Mirage2Matter}} supports efficient large-scale data synthesis, and VLA models trained on our generated data exhibit strong zero-shot generalization across various manipulation tasks. Our contributions are summarized as follows:

\begin{itemize}
\item \textbf{Faithful and photorealistic world representation.} 
We reconstruct high-fidelity visual representations of both target environments and manipulable objects using only ordinary multi-view videos.

\item \textbf{Physically grounded and robot-centric world modeling.} 
We leverage generative models to efficiently produce collision-ready geometric representations and introduce a calibration pipeline for accurate scale alignment, ensuring that the simulation remains consistent with the real world.

\item \textbf{Practical and scalable data generation for embodied learning.} 
Our approach builds efficient and scalable world models to support large-scale data simulation, enabling VLA models trained on the resulting data to achieve strong zero-shot generalization across diverse manipulation tasks.

\end{itemize}

\section{Related Work}
From a methodological perspective, existing simulated world models can be grouped into
reconstruction-based and generative-based approaches. While both paradigms aim to provide scalable data for embodied learning, they differ fundamentally in their assumptions, objectives, and system design. Our work adopts a reconstruction-based formulation, and we analyze how it compares to prior alternatives in terms of fidelity, scalability, and applicability.

\begin{figure*}
    \centering
    \includegraphics[width=\linewidth]{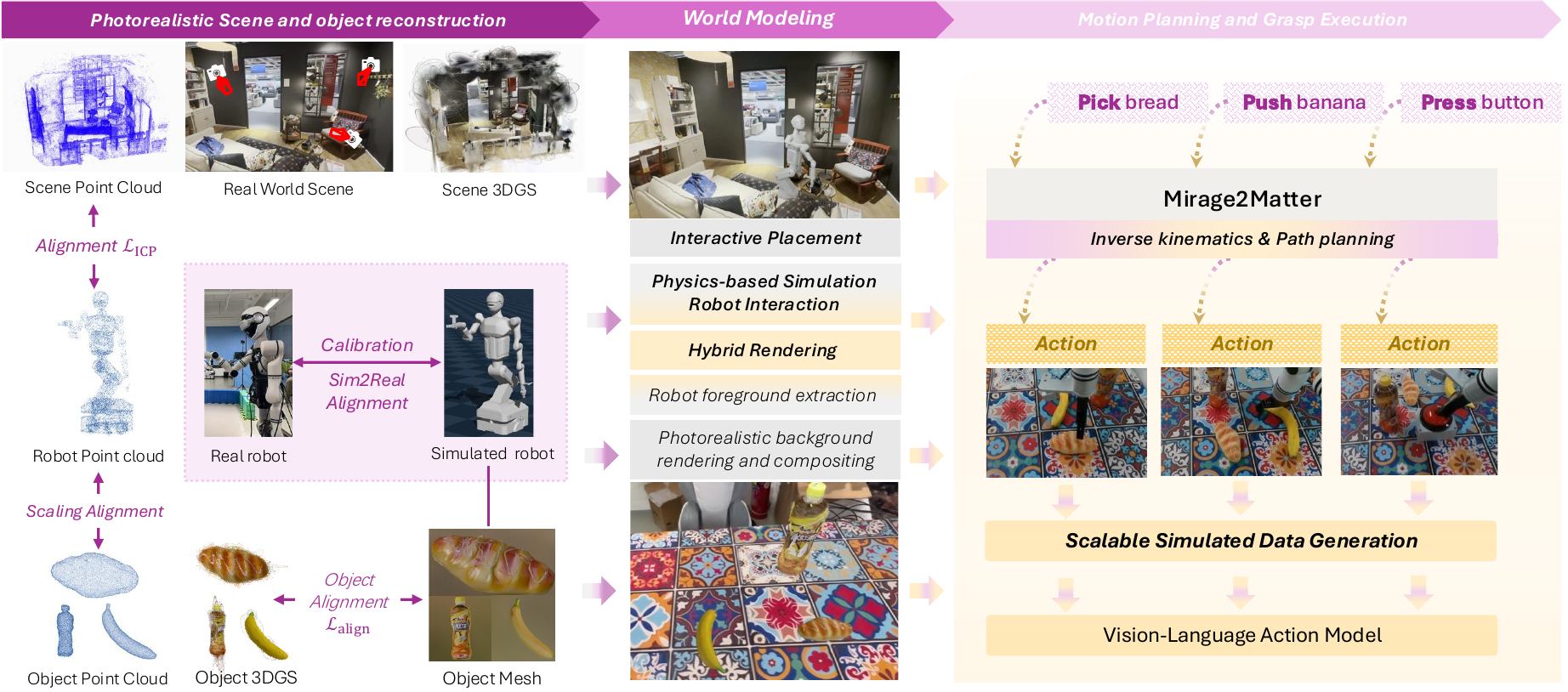}
    \caption{Overview of the Mirage2Matter framework. We reconstruct photorealistic scenes and objects from multi-view videos using 3DGS, align them to a robot-centric frame, and compose a unified world model for physics-based interaction. Hybrid rendering and motion planning are then used to generate scalable visual data for Vision–Language–Action training.}
    \label{fig:framework}
\end{figure*}
\subsection{Reconstruction-based World Models}
A range of systems leverage 3DGS to enhance visual fidelity in physics-based
simulators, enabling large-scale policy learning and evaluation.
Frameworks such as DISCOVERSE \cite{discoverse}, 3DGSim \cite{3dgsim}, PolaRis , Orbit \cite{mittal2023orbit}, \cite{polaris} and GSWorld \cite{gsworld} focus on building modular, efficient simulation platforms that integrate neural rendering with physics engines \cite{todorov2012mujoco,Genesis,NVIDIA_Isaac_Sim,coumans2016pybullet} to support closed-loop policy training, sim-to-real transfer, and benchmarking of robotic manipulation policies. Similarly, Re$^{3}$Sim \cite{re3sim}, Splatsim \cite{splatsim} and RoboSimGS \cite{robogsim} reconstruct task-specific tabletop environments and employ hybrid representations to generate data for zero-shot or closed-loop policy learning, incorporating scripted data generation or articulated object modeling. Despite improved realism, these systems require precise calibration
or specialized sensing and are tightly coupled to specific robot embodiments,
which limits scalability and reuse across environments.


\subsection{Generative-based World Models}
In parallel, a growing line of work explores generative-based world models that synthesize embodied interaction data through learned video or 3D generation \cite{zhang2024clay}.
Representative approaches such as DreamGen~\cite{dreamgen}, Ctrl-World~\cite{ctrlworld},
EmbodiedGen~\cite{embodiedgen}, LucidSim~\cite{lucidsim}, PhysGen3D~\cite{physgen3d}, PhysSplat \cite{physsplat}
and OmniPhysGS~\cite{omniphysgs} leverage large-scale diffusion models or generative priors to imagine future interactions conditioned on language, actions, or physical prompts. These methods demonstrate impressive scalability and diversity, enabling data augmentation, policy evaluation, and even policy improvement without requiring extensive real-world rollouts.

Despite their strengths, generative-based approaches are not designed to faithfully replicate specific real-world environments. Instead, they prioritize diversity and controllability over environment-specific geometric accuracy, often synthesizing scenes in an implicit or prompt-driven manner. As a result, the generated worlds may deviate from the exact layout, object geometry, lighting conditions, and camera configurations of a target deployment environment. Moreover, many of these systems operate purely in the image or latent space, relying on learned or approximate physics, which limits their ability to guarantee precise physical consistency
and metric alignment with real robotic platforms. These characteristics make generative world models well suited for large-scale data augmentation and imagination-based reasoning, but less applicable when faithful digital twins of real environments are required.

\section{Method}

We propose \textbf{Mirage2Matter}, a reconstruction-based world model that couples a \emph{photorealistic} world model with a \emph{physics} simulator, a brief overview is shown in Fig. \ref{fig:framework}. In this section we introduce the three parts of our work: $i.e.,$ scene and object reconstruction, cross-domain alignment, and data generation.
\subsection{Photorealistic Scene and Object Reconstruction}
\label{subsec:reconstruction}
This stage produces two complementary asset types:
(1) photorealistic 3DGS models for rendering (scene and objects), and
(2) explicit object meshes for physical simulation.
We therefore implement two branches:
a reconstruction branch(COLMAP+3DGS) and a generation branch (Tripo3D meshes \cite{tripo3d}).

\label{subsec:3dgs}
\paragraph{\textbf{3D Gaussian Splatting}} We utilize 3DGS~\cite{3DGS} as our photorealistic representation.
We model a scene as a set of $N$ anisotropic 3D Gaussian primitives, which can be efficiently rendered from novel viewpoints via rasterization and splatting~\cite{rasterization,splatting}.
Each primitive $i$ is parameterized by its mean location $\boldsymbol{\mu}_i\in\mathbb{R}^3$, anisotropic scale (stored as log-scales) $\boldsymbol{\ell}_i\in\mathbb{R}^3$, rotation $\mathbf{R}_i\in\mathrm{SO}(3)$ (or quaternion $\mathbf{q}_i$), opacity $\alpha_i\in[0,1]$, and spherical-harmonics (SH) coefficients for view-dependent color.
The corresponding (unnormalized) Gaussian density at $\mathbf{x}\in\mathbb{R}^3$ is
\begin{equation}
\label{eq:gaussian_density}
G_i(\mathbf{x})
=
\exp\!\left(
-\frac{1}{2}(\mathbf{x}-\boldsymbol{\mu}_i)^{\top}\,\Sigma_i^{-1}\,(\mathbf{x}-\boldsymbol{\mu}_i)
\right),
\end{equation}
where the world-space covariance $\Sigma_i\in\mathbb{R}^{3\times 3}$ is constructed from rotation and axis-aligned scales as
\begin{equation}
\label{eq:covariance_param}
\Sigma_i
=
\mathbf{R}_i\;
\mathrm{diag}\!\big(\exp(\boldsymbol{\ell}_i)^2\big)\;
\mathbf{R}_i^{\top}.
\end{equation}
We denote the full set of Gaussian parameters as
\begin{equation}
\label{eq:gaussian_set}
G=\{(\boldsymbol{\mu}_i,\boldsymbol{\ell}_i,\mathbf{R}_i,\alpha_i,\mathrm{SH}_i)\}_{i=1}^{N}.
\end{equation}

\paragraph{\textbf{Scene reconstruction (photorealistic background).}}
We capture a short handheld video of the target environment $V=\{I_t\}_{t=1}^{n}$ with sufficient parallax.
Before capture, we place a \textbf{calibration board} at the intended robot base location (floor/tabletop).
The board provides a planar region with known physical size, which later enables reliable metric alignment to the simulator frame.
From $V$, we sample keyframes and run COLMAP \cite{COLMAP_SfM, COLMAP_mvs, vote_engine}, to obtain sparse SfM points $P_{\mathrm{SfM}}=\{\mathbf{p}^{\mathrm{SfM}}_i\}$ and camera-to-world poses $\{\mathbf{T}^{w\leftarrow c}_t\}$.
We then train a scene 3DGS using these poses.
We use a photometric objective of the form
\begin{equation}
\label{eq:photo_scene}
\mathcal{L}_{\mathrm{photo}}
=
\sum_{t}\;
\bigl\|\hat{I}_t - I_t\bigr\|_1,
\end{equation}
where $\hat{I}_t$ is the rendering of the current 3DGS from the camera pose at time $t$ (other standard terms such as SSIM can be added as in~\cite{3DGS}).
The resulting scene 3DGS captures both global layout and fine-grained texture.

\paragraph{\textbf{Object Reconstruction (photorealistic foreground assets).}}
To reconstruct each manipulable object, we segment object masks on selected frames using SAM2 guided by text prompts.
Let $M_t^{(o)}\in\{0,1\}^{H\times W}$ be the mask for object $o$ at frame $t$.
We optimize an object-specific 3DGS $G_o$ using only masked pixels, e.g.,
\begin{equation}
\label{eq:photo_obj}
\mathcal{L}^{(o)}_{\mathrm{photo}}
=
\sum_t
\bigl\|
M_t^{(o)}\odot \hat{I}_t^{(o)} - M_t^{(o)}\odot I_t
\bigr\|_1,
\end{equation}
which suppresses background clutter and yields an object-only Gaussian field suitable for later composition and alignment.

\paragraph{\textbf{Mesh Generation (physics-ready geometry).}}
While 3DGS is well suited for rendering, physical simulation typically requires explicit surface geometry for collision detection and contact.
For each object, we capture single-view photos from distinct angles and provide a text description to Tripo3D \cite{tripo3d}, which produces a watertight mesh $M_o$ in a canonical pose.
Any remaining scale/pose discrepancy between the object 3DGS $G_o$ and mesh $M_o$ is resolved by cross-domain alignment (\cref{subsec:alignment}).

\begin{figure}
    \centering
    \includegraphics[width=\linewidth]{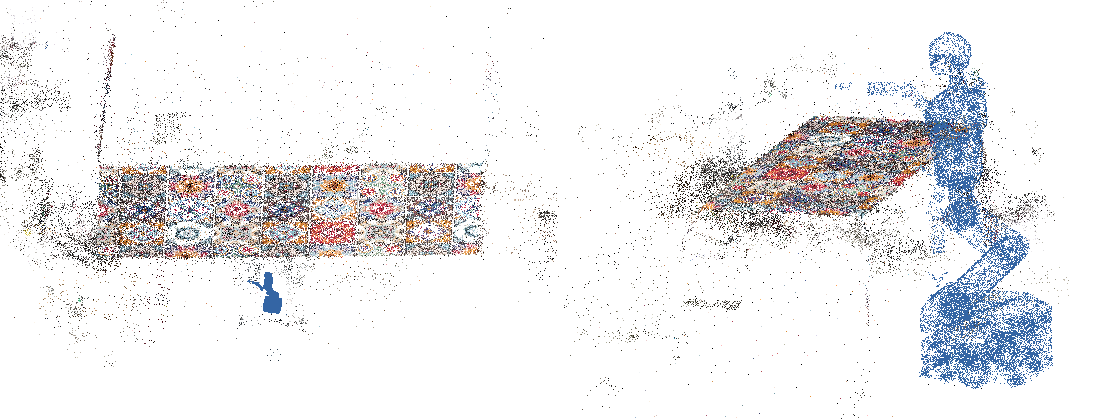}
    \caption{Visualization of scene–robot alignment. Left: unaligned scene point cloud (colored) and robot point cloud (blue). Right: aligned point clouds after scaled ICP between the calibration-board region and the robot base.}
    \label{fig:scene_alignment_comparison}
\end{figure}

\subsection{Cross-domain Alignment}
\label{subsec:alignment}
Reconstruction and mesh generation outputs obtained in section \ref{subsec:reconstruction} do not inherently share the simulator coordinate system.
We therefore align all assets to the Genesis \cite{Genesis} world frame, which we define as the \emph{robot base frame}.
We align geometry using similarity transforms in Eq.~\eqref{eq:icp}, while keeping camera and robot poses as rigid transforms in Eq.~\eqref{eq:obj_align}.

\paragraph{\textbf{Scene Alignment: SfM} $\rightarrow$ \textbf{Genesis (pre-align before 3DGS training).}}
We create a virtual robot workspace in Genesis, export it as a mesh, and sample it to a point cloud $P_r$.
From COLMAP we obtain the SfM points $P_{\mathrm{SfM}}$ and camera-to-world poses $\{\mathbf{T}^{w\leftarrow c}_t\}$.
Applying a similarity transform \emph{after} training the scene 3DGS can degrade rendering because it globally distorts learned Gaussian ellipsoids.
Instead, we estimate the similarity at the SfM level and train the scene 3DGS directly in the aligned frame.

We run scaled ICP between two semantically corresponding regions:
(i) the robot base region in $P_r$ and (ii) the calibration-board region in $P_{\mathrm{SfM}}$.
We estimate $\mathcal{S}=(s,\mathbf{R},\mathbf{t})$ by minimizing:
\begin{equation}
\label{eq:icp}
\mathcal{L}_{\mathrm{ICP}}
=
\sum_i \left\|\, s\,\mathbf{R}\mathbf{p}^{\mathrm{SfM}}_i+\mathbf{t}-\mathbf{p}^{r}_i \right\|_2^2.
\end{equation}
We transform SfM points as $\tilde{\mathbf{p}}^{\mathrm{SfM}}_i = s\,\mathbf{R}\mathbf{p}^{\mathrm{SfM}}_i+\mathbf{t}$.
For camera poses, we keep the pose itself while updating its translation consistently with the global similarity.
Let $\mathbf{T}^{w\leftarrow c}_t=[\mathbf{R}^{w\leftarrow c}_t\;\;\mathbf{t}^{w\leftarrow c}_t;\; \mathbf{0}^\top\;\;1]$.
Under $\mathcal{S}$, the aligned camera-to-world pose is
\begin{equation}
\label{eq:pose_update}
\tilde{\mathbf{R}}^{w\leftarrow c}_t = \mathbf{R}\,\mathbf{R}^{w\leftarrow c}_t,
\qquad
\tilde{\mathbf{t}}^{w\leftarrow c}_t = s\,\mathbf{R}\,\mathbf{t}^{w\leftarrow c}_t + \mathbf{t}.
\end{equation}
We then train the scene 3DGS using the transformed SfM points and camera poses, yielding a model that is natively aligned to Genesis robot base as shown in Fig. \ref{fig:scene_alignment_comparison}.
\begin{figure}
    \centering
    \includegraphics[width=\linewidth]{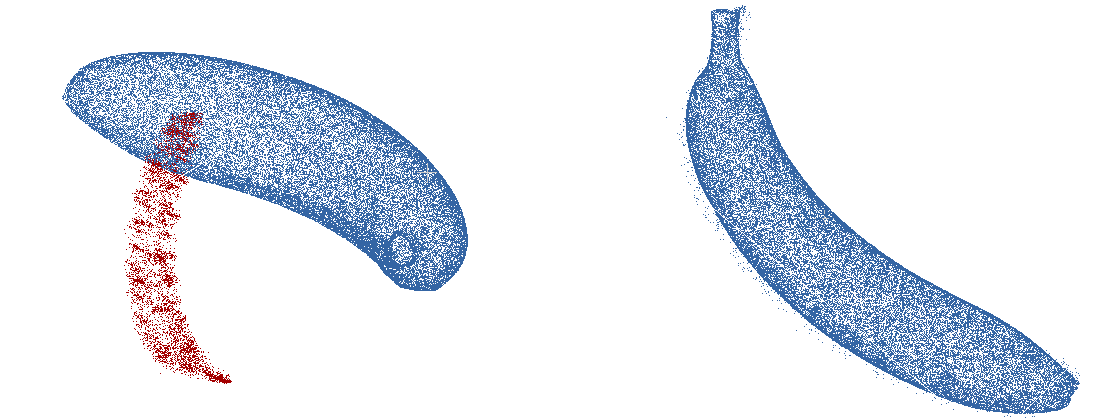}
    \caption{Object-level alignment between 3DGS and mesh representations. Blue points denote the mesh converted to a point cloud, and red points denote the corresponding 3DGS point cloud. Left:  object representations. Right: aligned result after similarity-based ICP.}
    \label{fig:object_alignment_comparison}
\end{figure}
\paragraph{\textbf{Object Alignment: Object 3DGS }$\leftrightarrow$ \textbf{Mesh (post-align)}.}
For each object, we align its object 3DGS $G_o$ to the corresponding mesh $M_o$.
We sample point clouds $P_o^{g}$ from $G_o$ and $P_o^{m}$ from $M_o$.
To obtain a robust initialization, we select a small set of corresponding keypoints $\{(\mathbf{p}^g_i,\mathbf{p}^m_i)\}$ across the two point clouds, then solve for an initial similarity $\mathcal{S}_o=(s_o,\mathbf{R}_o,\mathbf{t}_o)$ by:
\begin{equation}
\label{eq:obj_align}
\mathcal{L}_{\mathrm{align}}
=
\sum_i \left\|\, s_o\,\mathbf{R}_o\mathbf{p}^g_i+\mathbf{t}_o-\mathbf{p}^m_i \right\|_2^2,
\end{equation}
followed by ICP refinement. The visualization of this alignment is shown in Fig. \ref{fig:object_alignment_comparison}.
Unlike the scene, we apply this alignment \emph{after} object 3DGS optimization; empirically this has negligible impact on object rendering quality while simplifying the pipeline.

\paragraph{\textbf{Applying the Transformation to Object 3DGS Parameters.}}
Given $\mathcal{S}_o=(s_o,\mathbf{R}_o,\mathbf{t}_o)$ obtained from Eq. \ref{eq:obj_align}, we transform each Gaussian primitive in the object 3DGS.
For the mean:
\begin{equation}
\label{eq:mu_transform}
\tilde{\boldsymbol{\mu}}_i = s_o\,\mathbf{R}_o\boldsymbol{\mu}_i + \mathbf{t}_o.
\end{equation}
For the ellipsoid geometry, the induced covariance transform is
\begin{equation}
\label{eq:sigma_transform}
\tilde{\Sigma}_i = s_o^{2}\,\mathbf{R}_o\,\Sigma_i\,\mathbf{R}_o^{\top},
\end{equation}
which is equivalent to updating the internal parameterization as
\begin{equation}
\label{eq:internal_transform}
\tilde{\mathbf{R}}_i=\mathbf{R}_o\mathbf{R}_i,
\qquad
\tilde{\boldsymbol{\ell}}_i=\boldsymbol{\ell}_i + (\log s_o)\mathbf{1}.
\end{equation}
SH coefficients and opacity remain unchanged.
This yields an object 3DGS whose geometry is co-located with the physical mesh $M_o$, enabling unified rendering and simulation.

\paragraph{\textbf{Calibration and Sim-to-Real Alignment Strategy}}
\begin{figure}
    \centering
    \includegraphics[width=\linewidth]{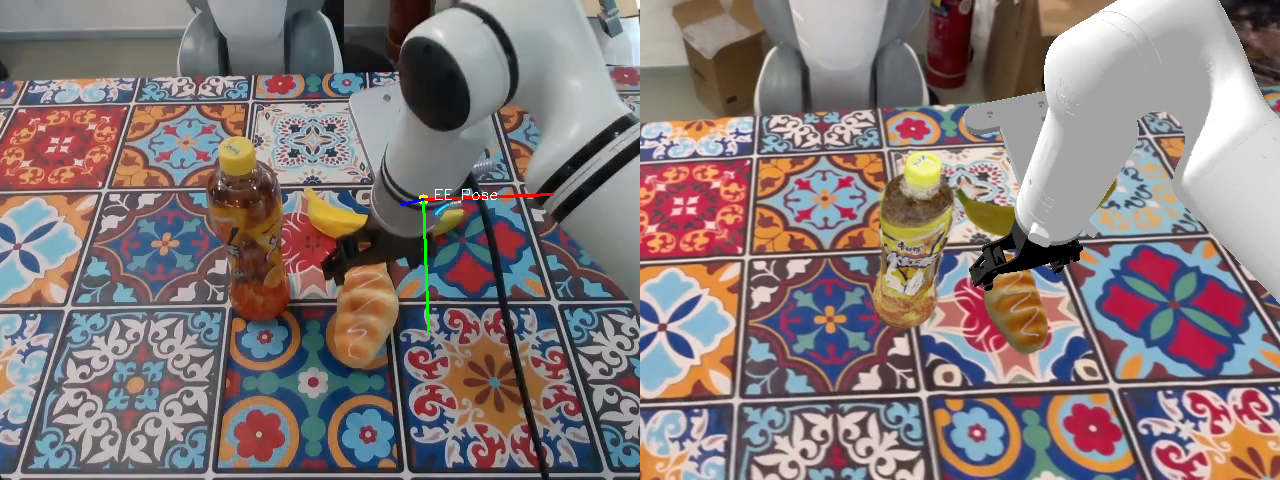}
    \caption{Sim-to-real egocentric camera alignment after calibration. Left: real head-mounted camera image. Right: simulated camera rendering. Under identical arm configurations, the manipulator appears at the same relative position in both views, demonstrating accurate camera alignment.}
    \label{fig:head_cam_calibration}
\end{figure}
To ensure geometric consistency between simulation and reality, we calibrate the robot's head camera relative to its manipulator base \cite{sim2real_important}. Using a ChArUco \cite{churaco} board attached to the end-effector, we collect paired data of camera detections and forward kinematics across randomized poses. We then solve the classical hand-eye calibration equation \cite{tsai_lenz} to obtain the extrinsic transformation. This transformation is integrated into the simulation's kinematic chain. In particular, when the manipulator appears within the field of view of the real camera, its projected position and relative pose in the simulated rendering remain consistent at the pixel level, as shown in Fig. \ref{fig:head_cam_calibration}.

\subsection{Data Generation}
\label{subsec:composition}
After alignment, all assets share the Genesis world frame.
We assemble a unified photorealistic 3DGS world which consists of scene and object 3DGS

\paragraph{\textbf{Interactive Placement}.}
For each object, SuperSplat provides an interactive placement transform
$\mathbf{T}_{\mathrm{place}}=(s_p,\mathbf{R}_p,\mathbf{t}_p)\in\mathrm{Sim}(3)$.
We record $\mathbf{T}_{\mathrm{place}}$ and apply it to the object's Gaussian parameters using the same update rules as in
Eqs.~\eqref{eq:mu_transform}--\eqref{eq:internal_transform} (with $s_o,\mathbf{R}_o,\mathbf{t}_o$ replaced by $s_p,\mathbf{R}_p,\mathbf{t}_p$.
This step yields semantically meaningful object layouts (e.g., placing a loaf of bread on a table) in a shared world frame.

\paragraph{Merging into a unified Gaussian world.}
We merge the scene and placed objects by concatenating their Gaussian parameter sets:
\begin{equation}
\label{eq:merge}
G_{\mathrm{world}}
=
G_{\mathrm{scene}}
\;\cup\;
\bigcup_{k=1}^{K} G_{\mathrm{obj},k}^{(\mathrm{placed})}.
\end{equation}
Because all assets are already aligned, no further joint optimization is required; $G_{\mathrm{world}}$ can be rendered directly from arbitrary viewpoints.

\paragraph{\textbf{Physics-based Simulation and Robot Interaction}}
\label{subsec:genesis}
We instantiate a physically grounded counterpart of the composed world in Genesis.
We load the robot from its URDF and define the robot base as the simulator world origin.
Since all alignment is performed in this frame as introduced in section \ref{subsec:alignment}, the robot requires no additional transform. For each object, we load its mesh $M_o$ and place it using the \emph{same} $\mathbf{T}_{\mathrm{place}}=(s_p,\mathbf{R}_p,\mathbf{t}_p)$ recorded in SuperSplat.
Concretely, for any mesh vertex $\mathbf{x}$ we apply the lifted transform:
\begin{equation}
\label{eq:mesh_place}
\mathbf{x}' = s_p\,\mathbf{R}_p\mathbf{x} + \mathbf{t}_p,
\end{equation}
ensuring that each mesh coincides with its corresponding Gaussian cluster in $G_{\mathrm{world}}$.
Genesis then provides collision checking, contact response, and physics-consistent motion for robot interaction with these mesh objects.

\paragraph{\textbf{Motion Planning and Grasp Execution}}
\label{subsec:planning}
We execute manipulation in Genesis using standard motion planning.
Given an object pose, we define a target end-effector pose $\mathbf{T}_{\mathrm{target}}$.
We solve inverse kinematics (IK) to obtain a target joint configuration $\mathbf{q}^{*}$ and plan a collision-free path using OMPL~\cite{ompl}.

\paragraph{Inverse kinematics.}
Let $\mathbf{T}_{\mathrm{ee}}(\mathbf{q})$ denote the forward-kinematics end-effector pose under joint angles $\mathbf{q}$.
We solve
\begin{equation}
\label{eq:ik}
\mathbf{q}^{*}
=
\arg\min_{\mathbf{q}}
\left\|
\mathbf{T}_{\mathrm{ee}}(\mathbf{q})
-
\mathbf{T}_{\mathrm{target}}
\right\|^2,
\end{equation}
where the pose error can be implemented as a weighted sum of position and orientation error (e.g., via a rotation-matrix or quaternion distance).

\paragraph{Path planning.}
Starting from the current joint configuration $\mathbf{q}_0$, we use RRT-based planning in joint space~\cite{plan_path} to obtain a collision-free trajectory
\begin{equation}
\label{eq:traj}
\mathcal{P}=\{\mathbf{q}_t\}_{t=0}^{T},
\end{equation}
which is then executed in Genesis to produce physically consistent robot motion and interactions.

\begin{table*}[t]
\centering
\caption{\textbf{Real-robot grasping success rate (\%) on two objects.}
All policies are trained using PoSA-VLA and evaluated on the real robot
with 30 trials per object.
\textbf{Real-World} is trained using real-robot demonstrations
and serves as a reference upper bound,
while other methods are trained purely in simulation
(300 demonstrations per object).}
\label{tab:grasp_main}
\begin{tabular}{lcccc}
\toprule
\textbf{Object}
& \textbf{Real-World}
& \textbf{DISCOVERSE}
& \textbf{RoboSimGS}
& \textbf{Mirage2Matter (Ours)} \\
\midrule
Banana     & 96.7 & 60.0 & 76.7 & \textbf{80.0} \\
Croissant & 90.0 & 66.7 & 76.7 & \textbf{86.7} \\
\bottomrule
\end{tabular}
\end{table*}
\paragraph{\textbf{Hybrid Rendering for Task Execution}}
\label{subsec:hybrid}
We generate training videos by combining physically correct robot motion (Genesis) with photorealistic rendering (3DGS).

\paragraph{Robot foreground extraction.}
We record a video in Genesis from the robot camera while executing $\mathcal{P}$.
We then apply Genesis video segmentation to extract the robot mask $M_t^{\mathrm{robot}}$ and the robot-only RGB frame $I_t^{\mathrm{robot}}$ at each timestep.

\paragraph{Photorealistic background rendering and compositing.}
For each timestep, we render the 3DGS world model $G_{\mathrm{world}}$ using a virtual camera that matches the Genesis robot camera, producing $I_t^{\mathrm{3DGS}}$.
We form the final hybrid frame via alpha compositing:
\begin{equation}
\label{eq:composite}
I_t^{\mathrm{final}}
=
M_t^{\mathrm{robot}}\odot I_t^{\mathrm{robot}}
+
\bigl(1-M_t^{\mathrm{robot}}\bigr)\odot I_t^{\mathrm{3DGS}},
\end{equation}
where $\odot$ denotes element-wise multiplication.
The resulting sequence $\{I_t^{\mathrm{final}}\}_{t=0}^{T}$ preserves physical correctness from Genesis (kinematics/dynamics and collision constraints) and visual realism from 3DGS, enabling scalable generation of photorealistic robot-interaction data for VLA training.



\section{Experiments}
\label{sec:experiments}

We evaluate whether photorealistic yet physically grounded data generated by Mirage2Matter
improves simulation-to-real generalization for VLA model learning.
Our primary evaluation protocol is Sim2Real VLA training: we collect task demonstrations in simulation,
train a VLA model on the simulated dataset, and directly deploy the resulting policy on a real robot
without any real-world fine-tuning.
We compare against two representative real-to-sim simulation pipelines,
DISCOVERSE~\cite{discoverse} and RoboSimGS~\cite{robogsim},
under a controlled and matched data collection and training budget.
As an upper-bound reference, we also report Real-World training using demonstrations
collected on the physical robot.

\subsection{Tasks and Real-World Testbed}
\label{sec:tasks}

We consider three manipulation tasks that cover diverse contact patterns and interaction primitives:
(1) \textbf{Grasp}, which requires grasping an object ;
(2) \textbf{Press Button}, which involves precise contact interaction with a button-like interface;
and (3) \textbf{Push/Pull}, which requires sliding an object on a tabletop via planar contact.

All real-world evaluations are conducted on the same physical robot and in the same workspace
from which the environment videos were captured to construct the simulated world,
ensuring a one-to-one correspondence between the simulated and real environments.

For each task and each object instance, we execute 30 real-robot trials and report success rate.
A trial is marked successful if the task-specific goal condition is achieved within a fixed time horizon
(e.g., stable lift for grasp; successful button activation; reaching a target displacement for push/pull).
We use identical success criteria across all methods.

\subsection{Data Collection in Simulation and Reality}
\label{sec:data_collection}

\paragraph{\textbf{Simulated Demonstrations.}}
We collect simulated demonstrations for the same tasks and object instances for each simulator.
To ensure fair comparison, we match the data budget across simulators,
recording 300 trajectories per task and object.
All simulators use identical robot embodiment, control frequency,
action parameterization, and egocentric RGB observations.
Task initialization distributions and termination conditions
are also kept the same.

\begin{table}[t]
\setlength{\tabcolsep}{11pt} 
\centering
\caption{\textbf{Task-level comparison between Mirage2Matter and Real-World training.}
Success rate (\%) is reported over 30 real-robot trials per task.}
\label{tab:task_compare}
\begin{tabular}{lcc}
\toprule
\textbf{Task} & \textbf{Real-World} & \textbf{Mirage2Matter} \\
\midrule
Press Button      & 96.7 & 93.3 \\
Push/Pull Objects & 83.3 & 73.3 \\
\bottomrule
\end{tabular}
\vspace{-0.2cm}
\end{table}

\paragraph{\textbf{Real-World Demonstrations.}}
As a practical upper bound, we additionally collect real-world demonstrations
on the physical robot.
For each task and object instance, we record 50 trajectories
and train the same VLA model, which is evaluated using the identical
real-robot test protocol.

\paragraph{\textbf{Data Collection Efficiency.}}
In practice, an experienced human operator can collect
approximately one real-robot demonstration per minute,
subject to fatigue and execution errors.
In contrast, Mirage2Matter can generate a comparable number of demonstrations
per minute on a single NVIDIA RTX~4090 GPU.
Simulated data generation can further run continuously and be parallelized,
enabling substantially higher scalability and lower marginal cost
for large-scale VLA training.

\subsection{Baselines and Training Protocol}
\label{sec:baselines}

\paragraph{\textbf{Baselines.}}
We compare our Mirage2Matter against two representative alternatives:
DISCOVERSE~\cite{discoverse}, a modular simulator framework that integrates
physics simulation with 3DGS-based rendering,
and RoboSimGS~\cite{robogsim}, a real-to-sim pipeline that reconstructs tabletop scenes
and generates simulation data with hybrid 3DGS/mesh representations.
For both baselines, we follow their recommended setup to build the simulated scene
and collect demonstrations, while matching the same task definitions
and dataset sizes described above.




\paragraph{\textbf{VLA Model.}}
All methods train the same VLA backbone, PoSA-VLA~\cite{li2025posa}.
PoSA-VLA is chosen for its strong performance on manipulation tasks
and its training efficiency, enabling controlled and fair comparisons
across different simulation pipelines.
We additionally experimented with a larger VLA backbone ($\pi_{0.5}$ \cite{pi05_2025}, OpenVLA-OFT \cite{kim2024openvla,kim2025fine})
and observed similar performance trends.
Due to its substantially higher training and inference cost,
we use PoSA-VLA as the default backbone for the majority of experiments.

\subsection{Implementation Details}
\label{sec:implementation_details}

\paragraph{\textbf{Robot Platform.}}
All real-world data collection and evaluation are conducted on the AlphaBot~1s robotic platform,
which is equipped with a 7-DoF manipulator, a head-mounted RGB camera,
and a wrist-mounted RGB camera.
Unless otherwise specified, the head-mounted camera is used as the primary egocentric visual input
for both simulation data generation and real-world deployment,
ensuring consistent observation modalities across domains.

\paragraph{\textbf{Training and Inference Setup.}}
Training is performed on a machine with a single NVIDIA A100 GPU with a batch size of 16
for a total of 200k training steps.
All inference and real-robot deployment experiments are conducted on an NVIDIA RTX~4090 GPU.
Unless otherwise noted, all optimization settings and hyperparameters
follow the default configuration reported in the original PoSA-VLA paper~\cite{li2025posa}.

\subsection{Evaluation Metrics}
\label{sec:metrics}

We report Success Rate (\%), defined as the fraction of trials
that successfully achieve the task goal.
Results are aggregated per task and per object and reported as the mean over trials.

\subsection{Main Results: Sim2Real VLA Generalization}
\label{sec:main_results}

\begin{figure*}
    \centering
    \includegraphics[width=\linewidth]{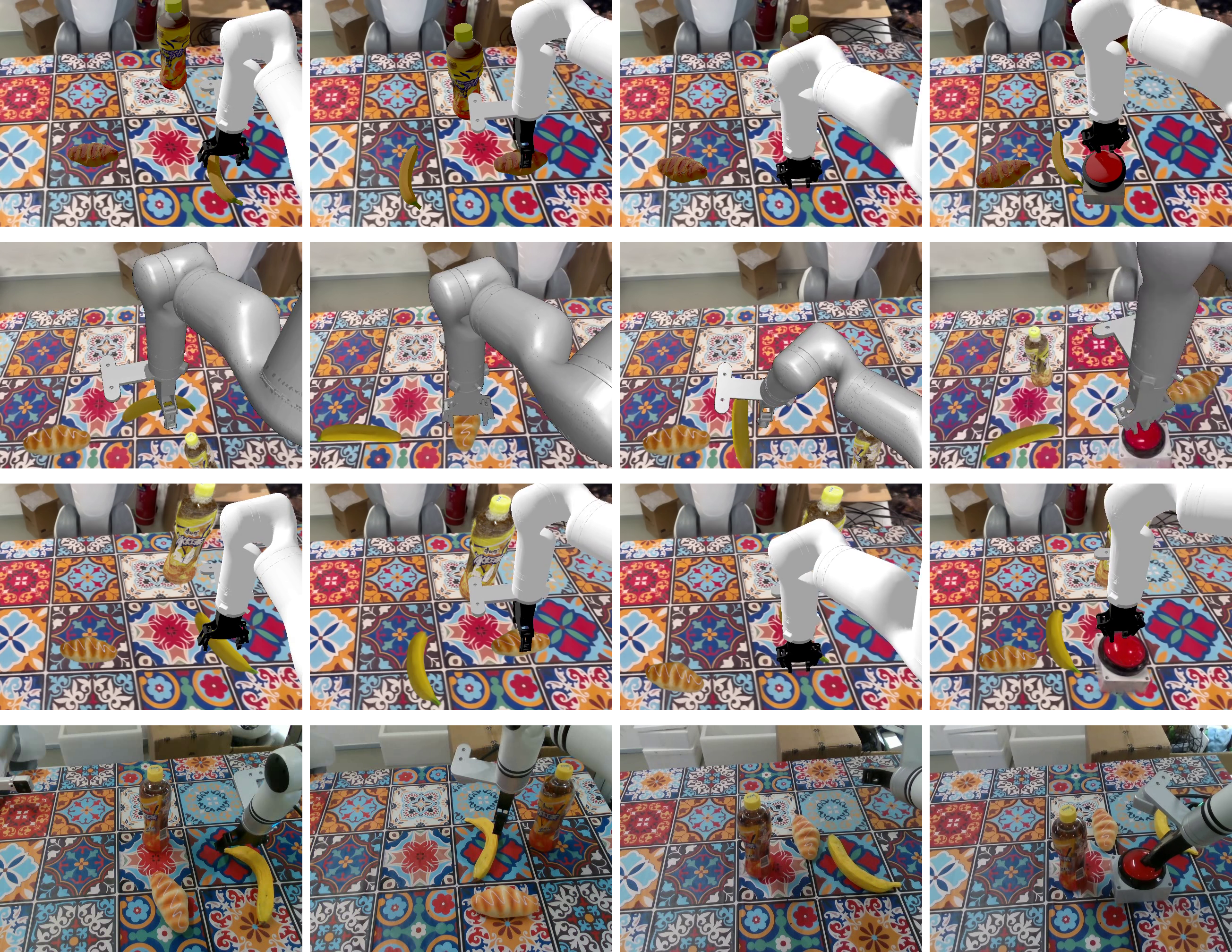}
    \caption{Qualitative comparison of different methods across multiple manipulation tasks. Each row corresponds to a specific simulator, from top to bottom: (1) RoboSimGS, (2) DISCOVERSE, (3) Ours, and (4) Real Robot. Each column represents a distinct task, from left to right: Grasp Banana, Grasp Bread, Push Banana, and Press Button. The images are keyframes extracted from videos recorded during task execution. Note that our method (third row) generates robust policies that successfully accomplish the manipulation goals in every scenario. In contrast to the baselines, our approach exhibits behavior that is both visually realistic and functionally effective, closely mirroring the successful execution of the real robot (bottom row), demonstrating superior fidelity compared to RoboSimGS and DISCOVERSE.}
    \label{fig:visualization_results}
\end{figure*}

\paragraph{\textbf{Grasping Performance on Real Objects.}}
Table~\ref{tab:grasp_main} reports real-robot grasp success rates on two everyday objects,
a banana and a croissant.
Compared with DISCOVERSE and RoboSimGS,
policies trained with Mirage2Matter achieve substantially higher success rates on both objects,
despite using the same number of simulated demonstrations.
This indicates that improving visual realism while preserving physically correct interaction
is critical for robust sim-to-real grasping, more visual results are shown in Fig. \ref{fig:visualization_results}.

\paragraph{\textbf{Task-level Comparison with Real-World Training.}}
Table~\ref{tab:task_compare} compares Mirage2Matter against Real-World training
across three manipulation tasks.
While Real-World training achieves the highest performance,
Mirage2Matter consistently attains strong success rates
and substantially narrows the performance gap,
demonstrating its effectiveness as a scalable alternative to real data collection. For more visual results, please refer to Fig. \ref{fig:visualization_results}.

\subsection{Ablation Studies}
\label{sec:ablation}

\paragraph{\textbf{Effect of Object Representation.}}
We study the impact of object representation by comparing
mesh-only objects with 3DGS-reconstructed objects within Mirage2Matter.
Table~\ref{tab:ablation_object} shows that replacing mesh objects with 3DGS objects
consistently improves real-robot grasping success,
highlighting the importance of visual consistency between objects and the surrounding environment.

\begin{table}[t]
\centering
\caption{\textbf{Ablation on Object Representation.}
Real-robot grasp success rate (\%) for Mirage2Matter
with and without 3DGS-based object reconstruction.
All results are obtained using PoSA-VLA.}
\label{tab:ablation_object}
\begin{tabular}{lccc}
\toprule
\textbf{Object Representation} & \textbf{Banana} & \textbf{Croissant} & \textbf{Average} \\
\midrule
w/o 3DGS & 76.7 & 80.0 & 78.4 \\
3DGS (ours)     & \textbf{80} & \textbf{86.7} & \textbf{83.4} \\
\bottomrule
\end{tabular}
\end{table}

\paragraph{\textbf{Effect of Training Data Scale.}}
We further analyze the effect of training data scale
by training PoSA-VLA with different numbers of simulated demonstrations
generated by Mirage2Matter.
As shown in Table~\ref{tab:data_scale},
performance improves consistently as the amount of training data increases,
with 300 demonstrations yielding the best real-world performance.

\begin{table}[t]
\setlength{\tabcolsep}{7pt} 

\centering
\caption{\textbf{Effect of training data scale.}
Real-robot grasp success rate (\%) as a function of the number of simulated demonstrations.}
\label{tab:data_scale}
\begin{tabular}{lccc}
\toprule
\textbf{\# Demonstrations} & \textbf{Banana} & \textbf{Croissant} & \textbf{Average} \\
\midrule
50  & 46.7 & 60.0 & 53.4 \\
150 & 76.7 & \textbf{86.7} & 81.7 \\
300 & \textbf{80.0} & \textbf{86.7} & \textbf{83.4} \\
\bottomrule
\end{tabular}
\vspace{-0.3cm}
\end{table}

\subsection{Analysis: Why Does Mirage2Matter Transfer Better?}
\label{sec:analysis}

We attribute the improved zero-shot real-world performance primarily
to the higher visual fidelity and stronger visual--physical consistency
of the observations generated by Mirage2Matter.
Compared with DISCOVERSE and RoboSimGS,
our pipeline produces egocentric views whose appearance statistics
(e.g., texture, lighting, and background geometry)
are significantly closer to those observed by the real robot,
thereby reducing the perceptual gap faced by the VLA model during deployment.

Importantly, Mirage2Matter maintains a consistent visual representation
for both the environment and manipulable objects
by rendering them within a unified 3DGS world.
In contrast, prior simulators typically combine mesh-based objects
with a 3DGS-rendered background,
introducing a systematic visual discrepancy between foreground objects
and their surrounding context.
Such inconsistency can inadvertently encourage shortcut learning,
where the model relies on simulator-specific visual artifacts
rather than learning interaction-relevant visual cues and action grounding.

At the same time, physical interactions in Mirage2Matter remain accurate
due to Genesis-based execution and collision-aware motion planning.
This prevents the model from overfitting to visually plausible
but physically incorrect behaviors,
and enables PoSA-VLA to learn robust perception--action mappings
that remain stable under real-world deployment.


    
        
        
        
        
        

\section{Discussion}

In this work, we present \textbf{Mirage2Matter}, a reality-consistent world model for low-cost, user-friendly data synthesis that enables zero-shot sim-to-real VLA training and supports scalable embodied intelligence. Our platform aligns 3DGS representations with collision-ready geometric proxies, yielding a physically grounded, high-fidelity visual simulation environment. It natively supports inserting new assets, editing existing assets within the reconstructed scene, and faithfully reproducing real-world lighting conditions, further reducing the visual gap between simulation and reality.
By training solely on our generated data, the VLM exhibits strong performance across diverse real-world downstream tasks.

In future work, we plan to support multiple simulation frames to make deployment easier and ensure compatibility across different projects.
Moreover, we will fully realize the potential of our method to build a large-scale interactive world and foster broader adoption in VLA training and evaluation by providing a fair, unified benchmark.
Beyond embodied AI, our method can further advance virtual reality (VR) by providing a more realistic experience of physical interaction.

\bibliographystyle{ACM-Reference-Format}
\bibliography{sample-base,software}
\end{document}